\documentclass[10pt,twocolumn]{article} 

\usepackage{graphicx}
\usepackage{array}
\usepackage{mdwmath}
\usepackage{mathtools}
\usepackage{amsmath}
\usepackage{breqn}
\usepackage{tabu}
\usepackage{url}

\usepackage[numbers,square]{natbib}
\usepackage[nohyperlinks, nolist]{acronym}

\usepackage{xcolor}
\usepackage[export]{adjustbox}

\usepackage{hyperref}

\definecolor{colorOfLink}{named}{black}

\hypersetup{%
	pdfencoding=auto,
	pdftitle={Automated Deep Photo Style Transfer},
	pdfauthor={Sebastian Penhou\"{e}t and Paul Sanzenbacher, University of T\"{u}bingen},
	pdfcreator={pdflatex - TeXstudio},
	pdfpagemode=UseOutlines,
	pdfdisplaydoctitle=true, 
	pdflang={English},
	colorlinks=true,
	linkcolor=colorOfLink,
	citecolor=colorOfLink,
	filecolor=colorOfLink, 
	menucolor=colorOfLink, 
	urlcolor=colorOfLink, 
	linktocpage=true, 
	bookmarksnumbered=true
}

\usepackage{bookmark} 

\usepackage{listings}
\lstMakeShortInline[columns=fixed]|

\usepackage{scrextend}
\usepackage{cleveref}
\usepackage{siunitx}
\usepackage{tabu}
\usepackage{caption}
\def\br#1{\left(#1\right)}

\def\Set#1{\left\{#1\right\}}

\title{Automated Deep Photo Style Transfer}

\usepackage{authblk}

\author{Sebastian Penhou\"{e}t} 
\author{Paul Sanzenbacher}
\affil{Department of Computer Science, University of T\"{u}bingen}

\begin{document}

\twocolumn[ 
	\begin{@twocolumnfalse} 

		\maketitle

		\begin{abstract}
			Photorealism is a complex concept that cannot easily be formulated mathematically.
			\emph{Deep Photo Style Transfer} is an attempt to transfer the style of a reference image to a content image while preserving its photorealism.
			This is achieved by introducing a constraint that prevents distortions in the content image and by applying the style transfer independently for semantically different parts of the images.
			In addition, an automated segmentation process is presented that consists of a neural network based segmentation method followed by a semantic grouping step.
			To further improve the results a measure for image aesthetics is used and elaborated.
			If the content and the style image are sufficiently similar, the result images look very realistic.
			With the automation of the image segmentation the pipeline becomes completely independent from any user interaction, which allows for new applications.
		\end{abstract}

		\vspace{0.35cm}

	\end{@twocolumnfalse} 
] 

\section{Introduction} \label{sec:intro}

Style transfer is a fast advancing technique that allows to transfer the visual style of one image to the content of another image.
\citeauthor{gatys16} created \emph{Neural Style Transfer} \cite{gatys16} which is a method for transferring the artistic style an image to another image by iteratively updating an initial noise image.
\emph{Deep Photo Style Transfer} \cite{luan17} is an approach by \citeauthor{luan17} that builds upon the technique created by \citeauthor{gatys16}.
It restricts the style transfer to changes in color space with respect to the images' context  and therefore allows a photorealistic style transfer.
This enables for enhancing images that were captured under suboptimal conditions with an image that has the perfect combination of color grading, lighting and contrast.
In addition, it allows texture changes in the image while avoiding significant distortions.
For example, in an image it is possible to change the daytime from day to night, the season from summer to winter and the weather from cloudy to sunny.

\begin{figure*}
	\includegraphics[width=\textwidth]{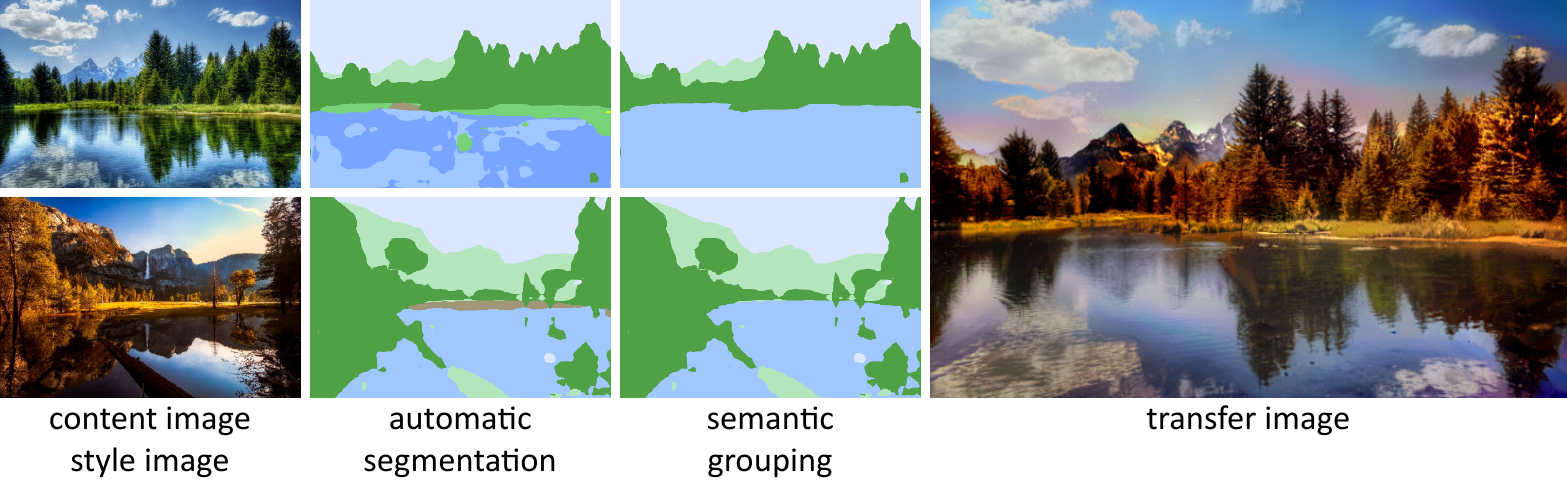}
	\caption{Given a content and style image, automatically a segmentation is created and semantically grouped to produce a transfer image in the size of the content image by using the \emph{Deep Photo Style Transfer} for 2000 iterations}
	\label{fig:teaser}
\end{figure*}

The approach presented in this paper builds upon \emph{Deep Photo Style Transfer}.
One of the main contributions of this paper is the automatic segmentation of input images and a semantic grouping thereof.
Another contribution of this paper is the optimization of the transfer image by improving the aesthetics of the image with the use of \emph{\acl{nima}} \cite{esfandari17}.
Results of these contributions can be seen in \Cref{fig:teaser}.
A \emph{TensorFlow} implementation of this paper is available on \emph{GitHub}\footnote{\emph{Automated Deep Photo Style Transfer} implementation on GitHub: \url{https://github.com/Spenhouet/automated-deep-photo-style-transfer}}.

In this paper, the automatic segmentation, the semantic grouping and the use of \emph{\acl{nima}} are elaborated in detail.

\section{Related Work} \label{sec:related_work}

There exists various work on transferring the style of a reference image to a target image, covering different purposes or addressing specific problems.
For instance, global color style transfer methods \cite{Reinhard2001,Pitie2005} aim to perform a color grading based on the reference image. These methods do however not support local color adaptations or texture changes.
In contrast to these purely analytical approaches, there are also machine learning approaches \cite{Laffont2014} and novel methods based on feature representations in neural networks \cite{Gardner2015}.
\citeauthor{gatys16} introduced an image style transfer algorithm based on a pre-trained \ac{cnn} \cite{simonyan14} that is originally used to classify images \cite{gatys16}.
Extending the approach by \citeauthor{gatys16}, \citeauthor{luan17} provide a method for enforcing photorealism on the result image \cite{luan17}.
This approach is also used as foundation for this work.
Recent work by \citeauthor{Li2018} introduces a novel algorithm for transferring the photo style of an image to another \cite{Li2018}.
This approach is an efficient closed-form solution consisting of a stylization and a smoothing step.

\section{Approach} \label{sec:style_transfer}

The \emph{Deep Photo Style Transfer} algorithm runs several pipeline steps that are necessary for computing the final transfer image.
This pipeline consists of creating a segmentation mask, grouping segmentation classes, defining and precomputing loss functions and gradually optimizing the transfer image.

First, the overall optimization process of the \emph{Neural Style Transfer} algorithm \cite{gatys16} is illustrated.
This approach is extended by augmenting the style loss by splitting it for the input images' segmentation.
Therefore, this loss is referred to as augmented style loss.
The \emph{Neural Style Transfer} algorithm does not preserve the photorealism of either of the input images.
To enforce photorealism on the transfer image and to preserve the details of the content image, an additional loss function called photorealism regularization is introduced.
In the original implementation of \emph{Deep Photo Style Transfer} \cite{luan17} the segmentation masks for the input images were created manually.
In addition, this work introduces an automated segmentation process that precomputes segmentation masks using another pre-trained neural network.
To further improve the appearance of the transfer image, an additional loss function, the image assessment loss, is proposed.
The image assessment loss is based on the \emph{Neural Image Assessment} \cite{esfandari17} and aims to improve the transfer images' aesthetics.

\subsection{Neural Style Transfer}

\begin{figure}
	\centering
	\includegraphics[width=\columnwidth]{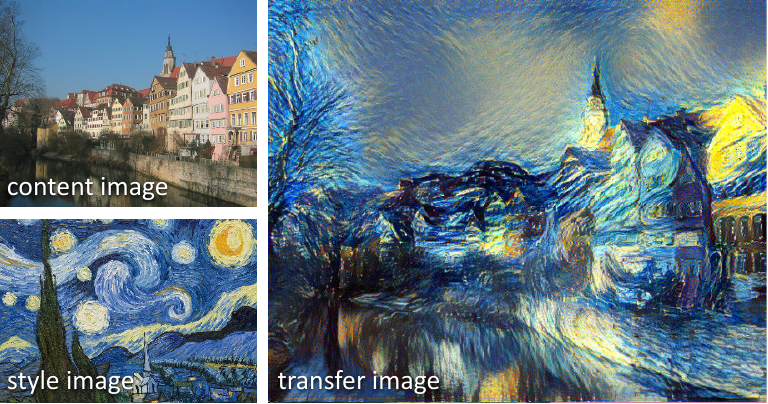}
	\caption{\emph{Neural Style Transfer} example of T{\"u}bingens Nackarfront in the style of Vincent van Gogh's \emph{The Starry Night}}
	\label{fig:style_transfer_example}
\end{figure}

\emph{Neural Style Transfer} \cite{gatys16} is a method for transferring the style of an image to another image using a pre-trained \ac{cnn}.
In this method, textures and patterns from the style image are transferred to the content image while maintaining distinctive details.
This form of style transfer is beneficial for showing an image in a specific artistic style.
\Cref{fig:style_transfer_example} shows an example where the artistic style of Vincent van Gogh's \emph{The Starry Night} is transferred to a photograph of the Neckarfront in T{\"u}bingen.
The transfer image has the same resolution as the content image.

This method uses a \ac{cnn} with 19 layers developed by the Visual Geometry Group of the University of Oxford \cite{simonyan14}, which is therefore called \emph{VGG19}.
The \emph{VGG19} network was originally trained on the \emph{ImageNet} dataset \cite{imagenet09} for image classification and object localization to investigate the effects of network depth.
Interestingly, the feature activations of different network layers of the \emph{VGG19} network can be used to extract style and content information from the input images.
Since the network is trained to classify all sorts of images, the feature activations encode various image textures and can thus be used to propagate texture information back to the input layer.
Given a content image and a reference style image, the texture information of the style image can be integrated into the content image by choosing an appropriate loss function.

\begin{figure*}
	\centering
	\includegraphics[width=\textwidth]{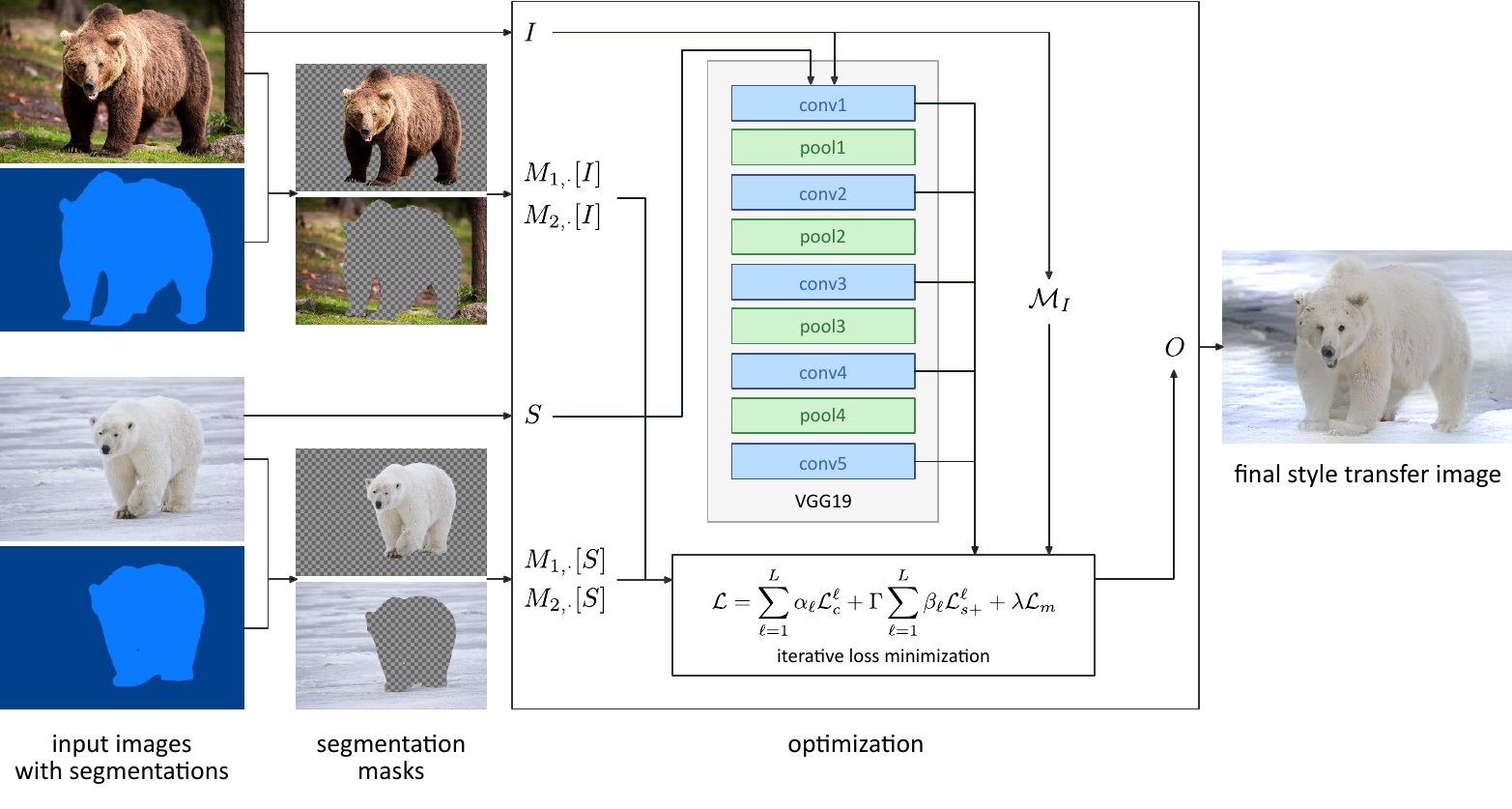}
	\caption{Overview over the optimization pipeline.}
	\label{fig:pipeline}
\end{figure*}

An overview over the optimization process described in this section is shown in \Cref{fig:pipeline}.
Given a content image $I$ and a style image $S$, the transfer image $O$ is computed by minimizing the loss function:
\begin{equation} \label{eq:style_transfer_loss}
	\mathcal{L}=\sum_{\ell=1}^L \alpha_\ell\mathcal{L}_c^\ell + \Gamma \sum_{\ell=1}^L\beta_\ell\mathcal{L}_s^\ell
\end{equation}
where $\ell$ denotes the convolutional layer index and $L$ the last convolutional layer in \emph{VGG19}.
$\Gamma$, $\alpha_\ell$ and $\beta_\ell$ for $\ell \in \Set{1,...,L}$ are constant weights that can be used to configure layer preferences or to weight the two loss functions.
For each convolutional layer, the content loss $\mathcal{L}_c^\ell$ is defined as:
\begin{equation}
	\mathcal{L}_c^\ell = \frac{1}{2N_\ell D_\ell}\sum_{ij}\br{F_\ell[O]-F_\ell[I]}_{ij}^2
\end{equation}
where for a given layer $\ell$, $N_\ell$ is the number of feature maps, $D_\ell$ the size of a feature map and $F_l$ the feature matrix accessed via indices $i$ and $j$.
The style loss $\mathcal{L}_s^\ell$ is defined as:
\begin{equation} \label{eq:style_loss}
	\mathcal{L}_s^\ell = \frac{1}{2N_\ell^2}\sum_{ij}\br{G_\ell[O]-G_\ell[S]}_{ij}^2
\end{equation}
where $G_\ell=F_\ell F_\ell^T$ is the Gram matrix that is defined as the inner product between the vectorized feature maps in layer $\ell$.

\subsection{Augmented Style Loss} \label{sec:augmented_style_loss}

In Neural Style Transfer, the style loss is computed over the entire image by encoding neural responses in the Gram matrices of the convolutional layers.
However, the style loss does not take semantic regions into account.
This results in textures being mapped into regions that do not semantically correspond to the texture, for example mapping the texture of a building onto the sky.
Using semantic segmentation of the input images, a separate style loss can be formulated for each semantic class in the segmentation.
A segmentation image has the same size as the original image and contains colored regions with each color representing a class.
For the augmented style loss, both the content image and the style image are split into binary segmentation masks scaled down multiple times for each convolutional layer $\ell$.
This  results in content segmentation masks $\mathcal{M}_{c,\ell}[I]$ and style segmentation masks $\mathcal{M}_{c,\ell}[S]$, where $c$ is the segmentation class and $\ell$ the convolutional layer index.

The augmented style loss $\mathcal{L}^\ell_{s+}$ then becomes:
\begin{equation}
	\mathcal{L}^\ell_{s+}=\sum_{c=1}^C \frac{1}{2N_{l,c}^2}\sum_{i,j}\br{G_{c,\ell}[O]-G_{c,\ell}[S]}_{ij}^2
\end{equation}
where the Gram matrices $G_{c,\ell}[O]$ and $G_{c,\ell}[S]$ are now computed using $F_{c,\ell}[O]=F_\ell[O]M_{c,\ell}[I]$ and $F_{c,\ell}[S]=F_\ell[S]M_{c,\ell}[S]$ with respect to the segmentation masks for class $c$.

\subsection{Photorealism Regularization}

Since the \emph{Neural Style Transfer} approach focuses on transferring the artistic style of an image onto another, there is no measure or parameter that can be used to control the photorealism of the result.
Furthermore, the transfer image in \emph{Neural Style Transfer} loses structures that are important for the transfer image to be perceived as photorealistic.
The aim of the photorealism regularization term is to extend the \emph{Neural Style Transfer} algorithm to yield photorealistic results.
However, characterizing photorealism mathematically is an unsolved problem \cite{luan17}.
Despite that, if the content image already is photorealistic, it is possible to maintain this property during optimization by preventing severe image distortions.
This can be achieved by creating a loss function based on locally affine transformations in color space that preserves edges in small patches.
Based on the work by \citeauthor{levin06} \cite{levin06}, a matting laplacian $\mathcal{M}_I$ is computed for the input image.
The matting laplacian is a closed-form image matting method to extract an object from an image \cite{luan17}.
This matting laplacian represents a gray scale matte as a locally affine combination of the \ac{rgb} input channels \cite{luan17}.

Let $V_c[O]$ be the vectorized color channel $c$ of the transfer image $O$. The affine loss is defined as:
\begin{equation}
	\mathcal{L}_m = \sum_{c=1}^3 V_c[O]^T\mathcal{M}_I V_c[O]
\end{equation}

\subsection{Optimization}

Combining \emph{Neural Style Transfer} with the augmented style loss $\mathcal{L}_{s+}^\ell$ and the photorealism regularization $\mathcal{L}_m$, the final optimization problem is given by minimizing the loss function:
\begin{equation}
	\mathcal{L}=\sum_{\ell=1}^L \alpha_\ell \mathcal{L}_c^\ell + \Gamma \sum_{\ell=1}^L \beta_\ell \mathcal{L}_{s+}^\ell + \lambda \mathcal{L}_m
\end{equation}
where, in addition to \Cref{eq:style_transfer_loss}, $\lambda$ is a weighting parameter for the affine loss $\mathcal{L}_m$.

The optimization problem is solved using the Adam optimizer \cite{diederik14} with a learning rate of $1.0$ as well as the parameters $\beta_1=0.9$, $\beta_2=0.999$ and $\varepsilon=10^{-8}$.
The transfer image $O$ can either be initialized using random noise, the content image or the style image.
Starting with the content image will most likely preserve structures and result in a slight color change.
Initializing with the style image will tend to produce results closer to the style but with major destruction of content structures, resulting in a loss of the photorealism property.
Using random noise as initialization yields a result balanced between the other two initialization options.

\subsection{Automatic Image Segmentation}

For the augmented style loss described in \Cref{sec:augmented_style_loss}, the style image and the content image need to be segmented into semantic groups.
This allows the algorithm to match semantically related regions in the input images and assign their corresponding textures correctly.
In the original paper \cite{luan17} there is no detailed information on how to create the segmentations from the input images.
However, the examples presented in the paper and on the released code\footnote{\emph{Deep Photo Style Transfer} GitHub project: \url{https://github.com/luanfujun/deep-photo-styletransfer}} indicate that they were created manually.
Since creating segmentations manually is tedious, an automatic segmentation algorithm is proposed in this section.

\paragraph{Neural Segmentation}

\begin{figure}
	\centering
	\includegraphics[width=\columnwidth]{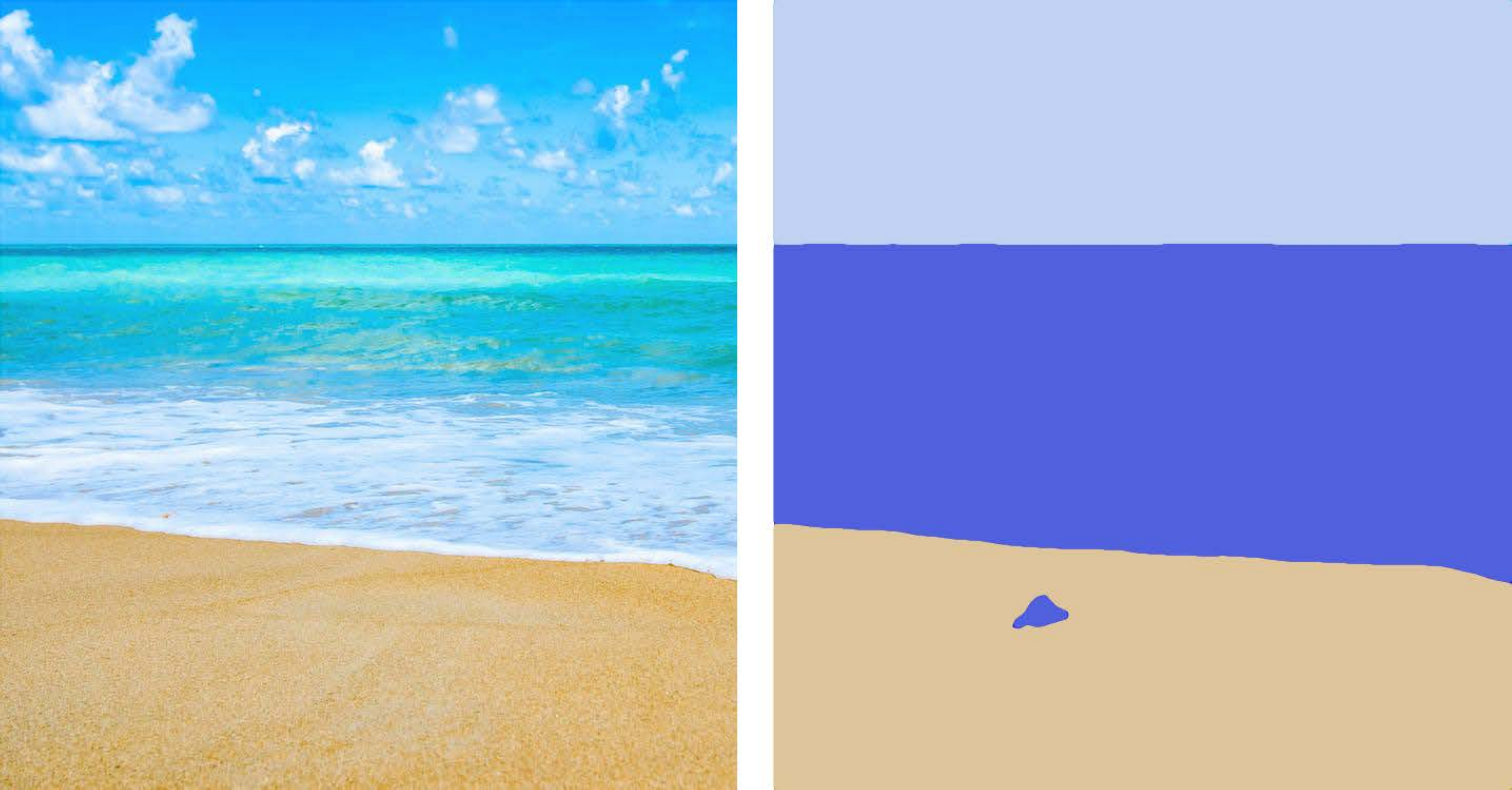}
	\caption{beach image with respective segmentation}
	\label{fig:segmentation_example}
\end{figure}

Segmentations are created with a pre-trained \ac{cnn} called \emph{\ac{pspnet}} \cite{zhao16} that was trained on the \emph{ADE20K} dataset \cite{zhou17}.
\emph{\ac{pspnet}} is ranked place 1 in the category Scene Parsing of the \emph{ImageNet} Challenge 2016\footnote{\emph{ImageNet} Challenge 2016: \url{http://image-net.org/challenges/LSVRC/2016/results}}.
This network creates a segmentation image as output that contains several colored regions representing content of one out of 150 classes.
\Cref{fig:segmentation_example} shows a beach image that is segmented into the classes sky, sea and sand by \emph{\ac{pspnet}}.

\paragraph{Semantic Grouping}

The \emph{Deep Photo Style Transfer} algorithm requires a small number of segmentation classes that cover large regions in the image to reduce memory usage and computation overhead.
However, depending on the input image, \emph{\ac{pspnet}} usually creates a segmentation image with 10 to 30 classes where many classes only cover small patches in the input image.
More importantly, closely related regions of the content and style image are not matched because they are assigned slightly different classes.
For example, river, lake and sea are three different classes and therefore result in differently colored segments.
To tackle these problems, a method for grouping segments of segmentation images based on their classes' semantic similarity is proposed.

Given an input image $I$ as a vector of $n$ pixels and a set $C$ of $n_C$ classes, the segmentation is a mapping $I\rightarrow C^n$.
Each class $l$ is represented by a set of words describing the class.

The semantic grouping for the set of classes from the content image $C_I$ and from the style image $C_S$, is done in two steps.
First, a difference merge is performed, followed bya class reduction based on a semantic threshold $\theta$.

The difference merge $d(C_A, C_B)$ replaces all class sets $D_A$ with a class set of $C_B$, where $D_A$ is the set difference $C_A - C_B$ for the images $A$ and $B$.
Every class set in $D_A$ is compared and replaced by the semantically most similar class set in $C_B$.
The semantic similarity between two class sets $sim_{\text{max}}$ is computed as the maximum semantic similarity $sim$ of all class pairs:

\begin{dmath}
	sim_{\text{max}}(l_1, l_2) = \text{max}\Set{sim(w_1, w_2): {w_1 \in l_1}, {w_2 \in l_2}}
\end{dmath}

The difference merge is first performed for the style image with $d(C_S, C_I)$ resulting in $C_S^*$ and then including this result for the content image with $d(C_I, C_S^*)$ resulting in $C_I^*$.
As a result, the segmentations for both images contain only the same class sets, yielding $C_I^* = C_S^*$.

The class reduction merges each pair of class sets of $C_{IS}^* = C_I^* \cap C_S^*$ that has a semantic similarity above the semantic threshold $\theta \in [0,1]$.
Every class sets can be interpreted as a graph $G$ with words representing a set of nodes $N$ and word pairs by a set of edges $E$.
Each edge in $E$ is assigned a corresponding semantic similarity of the connected nodes and the graph $G$ only contains edges with a semantic similarity $\textit{sim}$ above the semantic threshold $\theta$:

\begin{dmath}
	E = \Set{(w_1, w_2): {sim(w_1, w_2) > \theta} ,{w_1 \in l_1}, {w_2 \in l_2}}
\end{dmath}

To find the set $G^*$ of all connected subgraphs in $G$ a \ac{bfs} is performed.
The set of class sets from edges below the semantic similarity and the class sets contained in $G^*$ result in the set of merged class sets $C^*$.

The semantic similarity score $sim$ between two class words $w_1$ and $w_2$ is computed using a \ac{kg} from a Python library called \emph{Sematch}\footnote{\emph{Sematch} GitHub project: \url{https://github.com/gsi-upm/sematch}} \cite{zhu17}.
\emph{Sematch} provides a framework to evaluate semantic similarity by using one of multiple metrics.
In this work the similarity metric developed by \citeauthor{li03} \cite{li03} is used.

\begin{figure}
	\centering
	\includegraphics[width=\columnwidth]{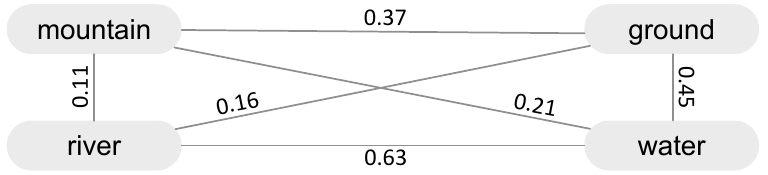}
	\caption{Semantic similarity graph with class words as nodes and semantic similarity scores on the edges}
	\label{fig:semantic_graph}
\end{figure}

An exemplary semantic similarity graph for the class words mountain, ground, river and water with the respective similarity scores is shown in \Cref{fig:semantic_graph}.

\begin{figure*}
	\includegraphics[width=\textwidth]{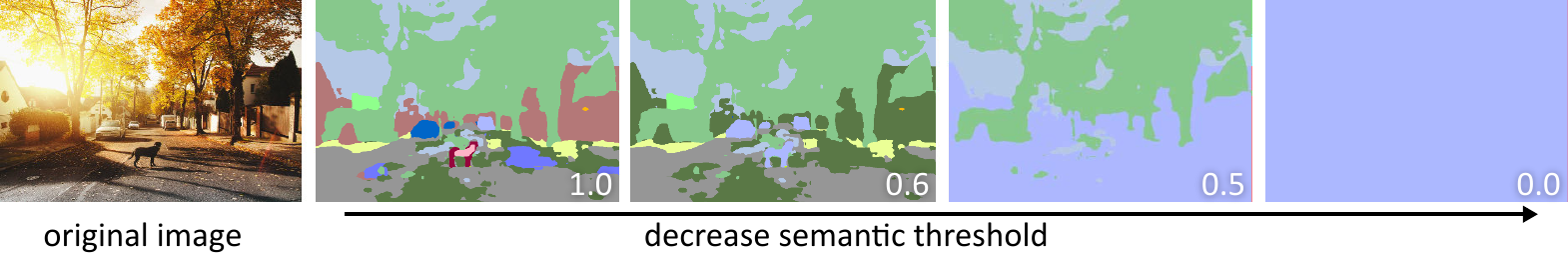}
	\caption{Effects of semantic grouping: With a decreasing threshold, semantically similar classes are merged.}
	\label{fig:merge_example}
\end{figure*}

The gradual effects of decreasing the semantic threshold for the semantic grouping are shown in \Cref{fig:merge_example}.

\subsection{Image Assessment Loss}

One part of achieving a photorealistic transfer image is to ensure image quality.
While this work focuses on objective image quality aspects like noise, artifacts and edge preservation, the assessment of image quality is also a very subjective task.
The aesthetics of an image for instance are subjectively perceived.
\citeauthor{esfandari17} developed \emph{\ac{nima}}, a model trained to rate the aesthetics of images based on human reference ratings \cite{esfandari17}.
As training data they used the \emph{\ac{ava}} dataset  that contains $255,500$ images of which each is rated by 200 people on average \cite{murray12}.
Every image is rated on its aesthetics with a rating between 1 and 10, with 1 for a bad looking image and 10 for the very aesthetically looking image \cite{murray12}.
\emph{\ac{nima}} was trained to predict the histogram of ratings for one image by providing a likelihood for every possible rating between 1 and 10 \cite{esfandari17}.
The mean rating of this histogram then provides an overall score of the images' aesthetics, that, as \citeauthor{esfandari17} claim, is close to the human ratings \cite{esfandari17}.
While \citeauthor{esfandari17} tried different pre-trained \ac{cnn} architectures for this task, this work focuses on the \emph{Inception-v2} architecture \cite{szegedy15}.

In \cite{talebi17} a neural network is trained to enhance images based on existing enhanced reference images.
To further improve the results, the assessed mean rating by \emph{\ac{nima}}, from here on just called \emph{\ac{nima}} score, is taken into the optimization with a very small factor of $0.0001$ \cite{talebi17}.

The image assessment loss introduced in this section is a similar approach to \cite{talebi17} and aims at improving the aesthetic of the transfer image.
The image assessment loss is defined as:

\begin{equation}
	\mathcal{L}_a = 10 - NIMA
\end{equation}

where $10$ is the best possible \emph{\ac{nima}} score and $\textit{NIMA}$ is the current \emph{\ac{nima}} score predicted by the \ac{cnn}.

Adding the image assessment loss $\mathcal{L}_a$ to \emph{Deep Photo Style Transfer}, the final optimization problem is given by minimizing the loss function:

\begin{equation}
	\mathcal{L}=\sum_{\ell=1}^L \alpha_\ell \mathcal{L}_c^\ell + \Gamma \sum_{\ell=1}^L \beta_\ell \mathcal{L}_{s+}^\ell + \lambda \mathcal{L}_m + \vartheta \mathcal{L}_a
\end{equation}

where $\vartheta$ is a weight to scale the image assessment loss.

\section{Results} \label{sec:results}

Findings about the transfer image initialization, the automatic segmentation, the optimization with \emph{\ac{nima}} and the computational performance aspect are examined in this section.
Example results are attached at the end of this document.

\subsection{Transfer Image Initialization}

\begin{figure*}
	\centering
	\includegraphics[width=\textwidth]{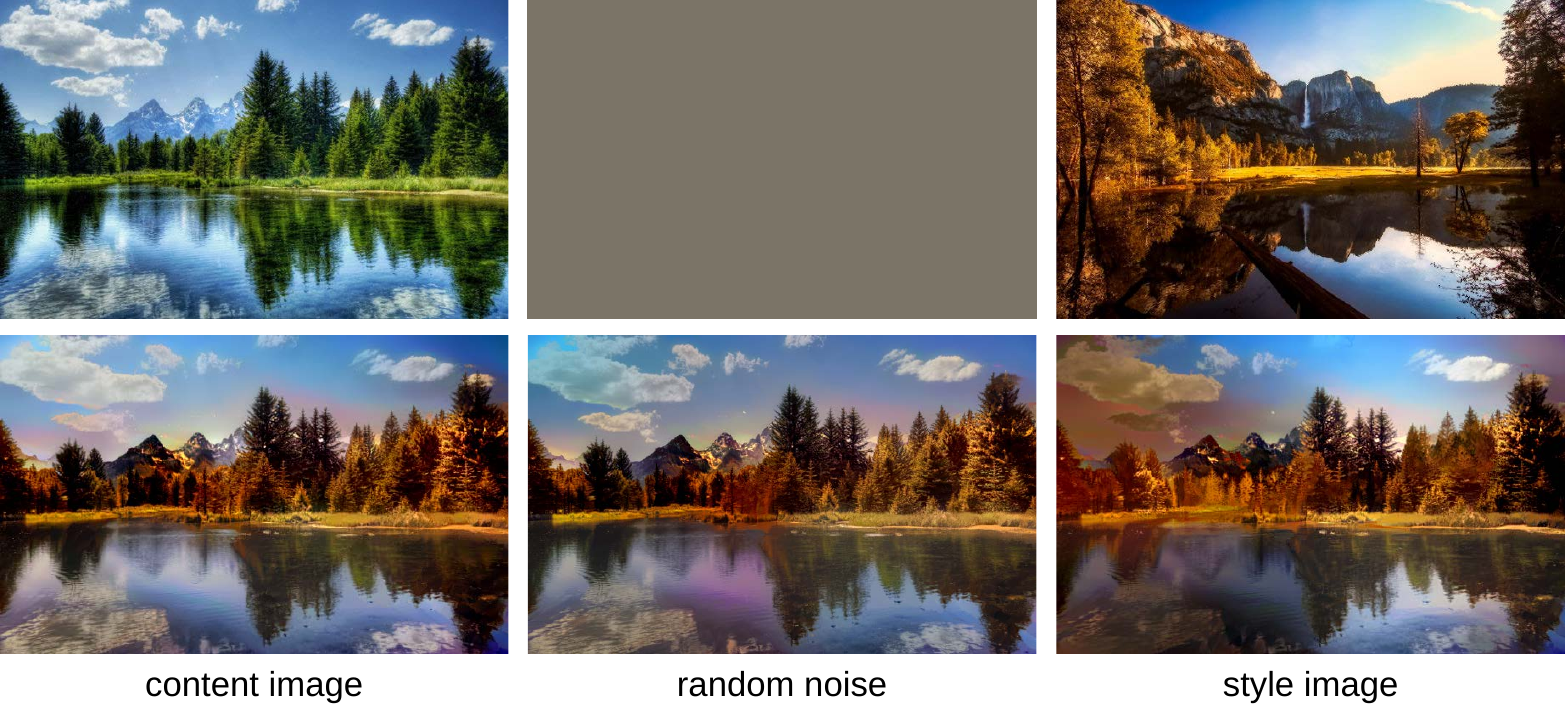}
	\caption{
		Results for different initializations of the transfer image.
		The top row shows the different initialization images, the bottom row shows the optimization results after 2000 iterations.}
	\label{fig:init_example}
\end{figure*}

In the original paper \cite{luan17}, random noise is used as an initialization of the transfer image.
On the one hand, using the content image as initialization can help to improve the photorealism. On the other hand, the optimization converges too fast.
In contrast, using random noise can help to integrate style patches in the transfer image that otherwise remain the same as in the content image.
Another option is to use images such as the style image, which does usually not converge to a photorealistic result.
A comparison of the different initialization options is shown in \Cref{fig:init_example}.

\subsection{Automatic Segmentation}

The automatic segmentation introduced in this paper reduces the external work that is necessary before running the \emph{Deep Photo Style Transfer}.
The need for manually created segmentation masks is completely removed.
Creating segmentation masks manually can consume an immense amount of time, especially when many different images are stylized.
While saving time can be a huge factor, manual segmentations allow for more goal-oriented segmentations and therefore more specific results.
The automatic segmentation implemented in this paper assumes that in order to achieve photorealism, the same classes in the images need to be mapped onto each other.
Although this is a reasonable assumption, it is not always the desired behavior.
Matching segmentation classes that are not semantically related can be used to create more abstract yet photorealistic results\footnote{Abstract examples can be seen in \cite{luan17}}.

It is worth mentioning that the used methods for the automatic segmentation, \emph{\ac{pspnet}} and \emph{Sematch}, introduce errors on their own.
The segmentation coverage and label accuracy might not always yield the excepted results.

\subsection{NIMA Optimization}

\begin{figure*}
	\includegraphics[width=\textwidth]{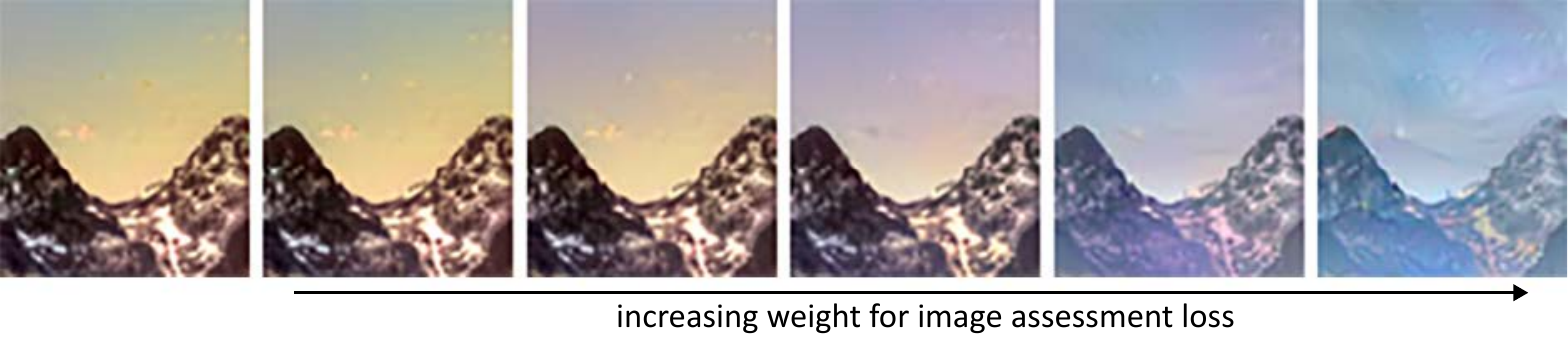}
	\caption{Effect of image assessment loss: With increasing weight, more artifacts appear }
	\label{fig:nima_example}
\end{figure*}

In \cite{talebi17} \citeauthor{talebi17} mention that weighting the \emph{\ac{nima}} score high leads to artifacts and low to barely any improvement.
With \emph{Deep Photo Style Transfer}, it was not possible to produce any notable improvement with various weights for the image assessment loss.
In contrast to that, a high weighting of the image assessment loss did lead to artifacts.
This can be seen in \Cref{fig:nima_example} where the first image is the result of a style transfer without the image assessment loss and the last image with a weight resulting in an image assessment loss four orders of magnitude as high as the other loss functions.
Over the training span the image assessment loss is reduced notably with a stable gradient while not resulting in sufficient improvements in image quality or aesthetics.

\subsection{Performance}

Empirical results show that in order for an image to transfer the style sufficiently, using content initialization, a minimum of 1000 iterations is needed.
Good results are achieved with about 2000 iterations on average.
After 4000 iterations there can be slight improvements, but most of the time they are negligible.

The following default parameters are recommended: semantic threshold $\theta=0.6$, content initialization, content weight $\alpha=1$, augmented style weight $\beta=100$, photorealism regularization weight $\lambda=10000$, image assessment weight $\vartheta=100000$ and $2000$ iterations.

Computing the matting laplacian is a very memory-intensive task.
This introduces a hard-upper limit on a combination of image size and segment count.
A computer with 8 GB of system memory hits this limit by an image size of maximum 700 pixels in width or height and with this resolution a segment count of 5.

On a computer with an \emph{Intel i7-3770K} processor, 8 GB of system memory, a \emph{NVIDIA GeForce GTX 1080} graphic card, \emph{Python} 3.6.4, \emph{TensorFlow} 1.4.0 and \emph{CUDA} 9.1.85 a regular stylization run takes about 16 minutes.

\section{Important Observations} \label{sec:observation}

While applying the \emph{Deep Photo Style Transfer} algorithm to various examples, some of the limitations of the approach were examined. The most important observations are described in this section.

\subsection{NIMA Gradient Information}

It is questionable if \emph{\ac{nima}} provides a meaningful gradient.
The \emph{\ac{nima}} score does not seem to provide a reasonable aesthetic rating.
It is possible that images that look worse subjectively than other images achieve a higher \emph{\ac{nima}} score.

In addition to the virtually absent results the application of the image assessment loss with random noise initialization can worsen the results in some cases.

Ignoring the above findings, it is questionable if the application of the \emph{\ac{nima}} score in the \emph{Deep Photo Style Transfer} setting makes sense.
A goal behind \emph{Deep Photo Style Transfer} is the transformation of style while maintaining the photorealism.
Making images look subjectively good or aesthetically does not necessarily correlate with images looking photorealistic.
In contrary, humans could judge an image as aesthetically pleasing while the image is far from or not photorealistic.

\subsection{ADE20K}

As already mentioned by \citeauthor{zhao16} in \cite{zhao16} the \emph{ADE20K} dataset contains some confusing classifications.
For the classes skyscraper, road and route the dataset contains two different segmentation entries, one for skyscraper and one for road and route, but assigns the same color to both segments.

Another problem is, that the classes ground, counter, screen and stool are all assigned to multiple colors.
Both problems result in major errors in the segmentation and semantic grouping process.
These problems were solved by merging duplicate colors and reducing duplicate classes.

\subsection{Sematch}

The \emph{ADE20K} dataset contains a lot of classes with two words for one class.
This is an issue since the \emph{Sematch} framework cannot handle classes that contain a space like ``computing device''.
Joining the words of a multi word class with an underscore solves this issue.

There are also words that \emph{Sematch} does not know, like ``arcade_machine''.
These words need to be substituted by a word with a similar concept, for example replacing ``arcade_machine'' with ``arcade''.

\subsection{Stylization Limitations}

While the style transfer in general show promising results it also shows deficiencies in specific cases.
In the case of reflective surfaces like water, windows and mirrors the content of the respective area can contain multiple objects at the same time.
For example, a window can reflect other objects, show objects behind it and can just be classified as a window.
Often the latter is the case, for example when an object and its reflection in the window will be stylized differently but should clearly look the same.
The problem here is that \emph{Deep Photo Style Transfer} does not provide any concept for mixed or multiple classes at one position.

Punctual light sources or bright spots in the content image that do not appear in the style image are another problematic case.
The style transfer seems to be unable to remove these spots and instead leaves an artifact like structure with a glowing border around it.
The problem seems to be that \emph{Deep Photo Style Transfer} does not have the information to replace an area completely and to introduce new edges and shapes instead of just recoloring the existing shapes.
Well-defined objects with well-known color and edge properties for specific parts are a problem as well.
The problem starts with \emph{\ac{pspnet}} being unable to provide such detailed and accurate segmentation masks to differentiate every single object.
\emph{Deep Photo Style Transfer} has no concept of when to apply a high precision coloring and when mixing up colors does not reduce the photorealism.

From a logical standpoint, these limitations break the photorealism in the transfer image.

\section{Future Work}

To produce a single style transfer image, it takes about 16 minutes on a \emph{NVIDIA GeForce GTX 1080}.
For the stylization of multiple images, this is considerably too long.
\citeauthor{Johnson2016} in \cite{Johnson2016} improved the time of \emph{Neural Style Transfer} for one style transfer by three orders of magnitude.
By pre-training the \emph{VGG19} with the style image they were able to transfer the style onto a content image in only one iteration.
An approach like this could also be used to improve the style transfer time of \emph{Deep Photo Style Transfer}.

The matting laplacian calculation is a limiting factor.
Replacing this calculation of the photorealism regularization is the key to running style transfer for high resolution images.
The replacement does not necessarily have to be a matting algorithm and can instead just be another function that achieves the same results.

The current segmentation algorithm, \emph{PSPNet}, only supports up to 150 classes.
These classes mainly are very broad categories.
It is impossible to segment finer details.
There is no concept of segments containing subsegments and thereby creating a hierarchy of classifications.
Such a concept, generally more classes and a higher accuracy would improve the segmentation.
If a future segmentation algorithm demonstrates better performance in these areas, it is an improvement to the method used in this work.

\emph{Deep Photo Style Transfer} aims to preserve the photorealism of the content image.
While this works considerably well, there is still room for improvement.
An interesting approach could be to train \acp{gan} \cite{Goodfellow2014} to discriminate and generate real life photographs.
The trained discriminator could be used to define a new photorealism loss.

\section{Conclusion}

We introduced an easy to use and time saving automatic segmentation as an extension to \emph{Deep Photo Style Transfer}.
The segmentation is complemented by an automatic semantic grouping algorithm that produces beneficial segmentations with reduced segments based on a semantic threshold.
Furthermore, we added an image assessment loss based on the \emph{\ac{nima}} model to improve image aesthetics.

Like the original implementation of \emph{Deep Photo Style Transfer}, our implementation is also limited by the resolution, the number of segmentation classes and execution time.

\begin{figure*}
	\centering
	\includegraphics[width=\textwidth]{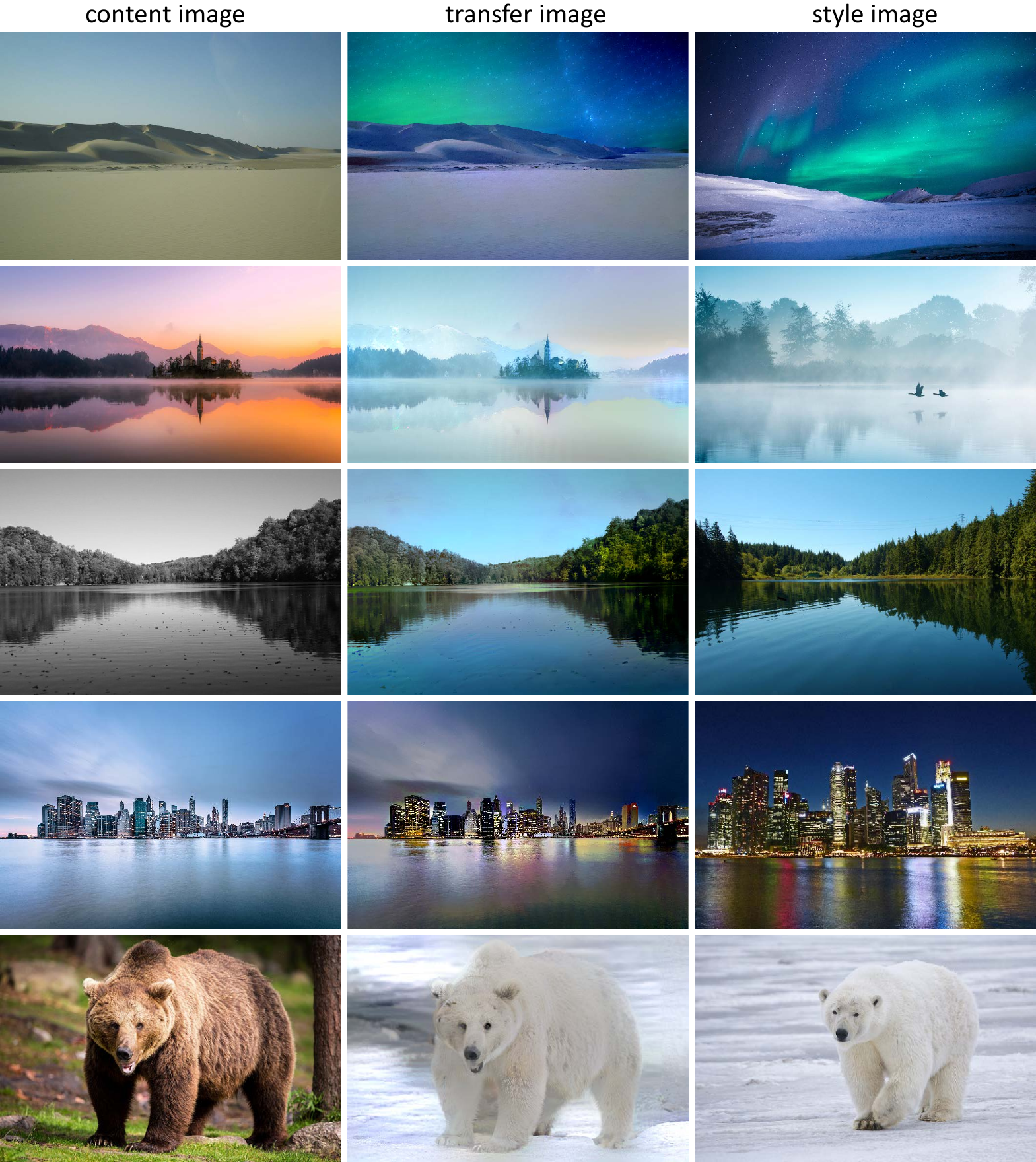}
	\caption{example results}
	\label{fig:nice_results}
\end{figure*}

\begin{figure*}
	\centering
	\includegraphics[width=\textwidth]{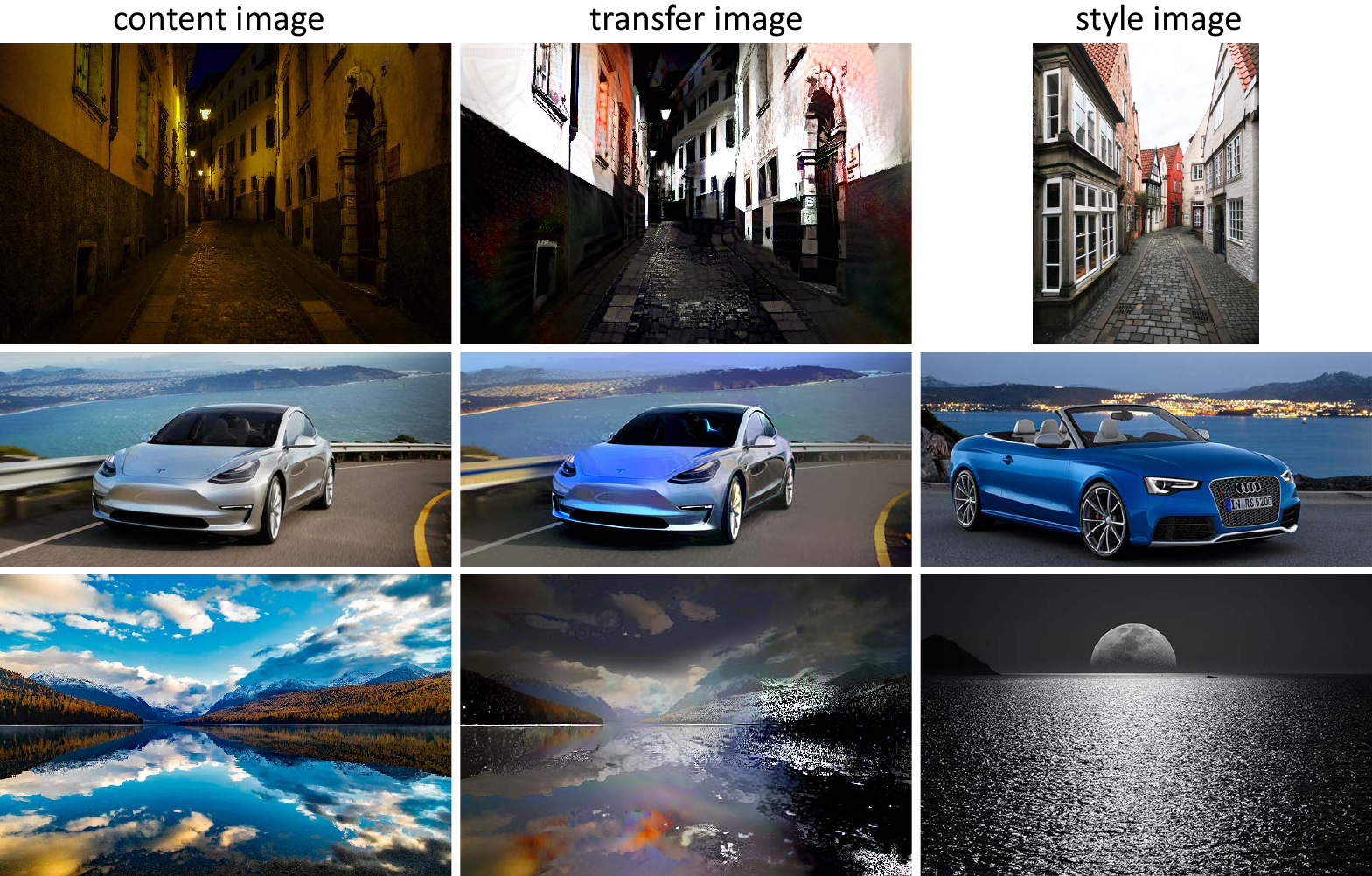}
	\caption{hallenging scenarios}
	\label{fig:challenging_scenarios}
\end{figure*}

However, the automated process ensures an easy usage and the results produced by our implementation are comparable to the results of the original implementation of \emph{Deep Photo Style Transfer}.
Example results are shown in \Cref{fig:nice_results} as well as some challenging scenarios in \Cref{fig:challenging_scenarios}.

\begin{acronym}
	\acro{ann}[ANN]{artificial neural network}
	\acro{cnn}[CNN]{convolutional neural network}
	\acro{rgb}[RGB]{red, green and blue}
	\acro{pspnet}[PSPNet]{Pyramid Scene Parsing Network}
	\acro{bfs}[BFS]{breadth first search}
	\acro{kg}[KG]{Knowledge Graph}
	\acro{ava}[AVA]{Aesthetic Visual Analysis}
	\acro{nima}[NIMA]{Neural Image Assessment}
	\acro{gan}[GAN]{generative adversarial network}
\end{acronym}

\bibliography{references}

\end{document}